\documentclass[conference, final, letterpaper, 10pt]{ieeeconf}
\IEEEoverridecommandlockouts
\usepackage{amsmath,amssymb,amsfonts}
\usepackage{algorithmic}
\usepackage{graphicx}
\usepackage{textcomp}
\usepackage{xcolor}
\usepackage{booktabs}
\usepackage{cite}
\usepackage{jabbrv}
\usepackage{url}

\def\BibTeX{{\rm B\kern-.05em{\sc i\kern-.025em b}\kern-.08em
    T\kern-.1667em\lower.7ex\hbox{E}\kern-.125emX}}

\title{A survey on robustness in trajectory prediction for autonomous vehicles}

\author{Jeroen Hagenus$^1$, Frederik Baymler Mathiesen$^2$, Julian F. Schumann$^1$, Arkady Zgonnikov$^1$
\thanks{$^1$Department of Cognitive Robotics, $^2$Delft Center for Systems and Control,
        TU Delft, 2628 CB Delft, The Netherlands
        {\tt\small \{J.Hagenus@student., F.B.Mathiesen@, J.F.Schumann@, A.Zgonnikov@\}tudelft.nl}}%
}

\begin{document}

\maketitle

\begin{abstract}
Autonomous vehicles rely on accurate trajectory prediction to inform decision-making processes related to navigation and collision avoidance. However, current trajectory prediction models show signs of overfitting, which may lead to unsafe or suboptimal behavior. To address these challenges, this paper presents a comprehensive framework that categorizes and assesses the definitions and strategies used in the literature on evaluating and improving the robustness of trajectory prediction models. This involves a detailed exploration of various approaches, including data slicing methods, perturbation techniques, model architecture changes, and post-training adjustments. In the literature, we see many promising methods for increasing robustness, which are necessary for safe and reliable autonomous driving.

\end{abstract}


\section{Introduction} \label{Introduction}
A critical technological aspect in the control of autonomous vehicles is forecasting future states of the surrounding environment to facilitate safe navigation through dynamic surroundings. The predicted future trajectories of traffic participants play an important role in detecting potential hazards in advance and are essential for the design of decision-making or planning algorithms \cite{huang_survey_2022, bharilya_machine_2024}. This predictive task relies on utilizing perceived data, such as camera and LiDAR data \cite{caesar_nuscenes_2020, ettinger_large_2021, chang_argoverse_2019,zhan_interaction_2019}. This data is mostly tracked across multiple timestamps, transformed into trajectories, and then integrated into maps to enable accurate predictions~\cite{caesar_nuscenes_2020,chang_argoverse_2019}.

The task of predicting future trajectories of traffic participants is a complex time series regression problem \cite{jiao_tae_2022}.
It involves intricate contextual elements, including directional information, velocity, position, and time. Furthermore, such trajectories inherently convey semantic and behavioral information, facilitating inference of the future movement of road users~\cite{jiao_tae_2022}.

Deep learning models have proven effective in trajectory prediction~\cite{yuan_agentformer_2021, salzmann_trajectron_2020, ngiam_scene_2022, liu_multimodal_2021,phan-minh_covernet_2020}. These models consider dynamic factors, model map information, and can incorporate interaction-related factors, allowing adaptation to complex scenes \cite{huang_survey_2022}. However, many deep learning models are known to be sensitive to small errors and susceptible to external attacks, potentially resulting in undesirable behavior and decreased performance \cite{xiong_towards_2022}. Therefore, ensuring reliable performance (often called \textit{robustness}) is key to the safe deployment of these deep learning models. In the context of trajectory prediction, robustness implies the ability to produce predictions unaffected by spurious features (e.g., sensor noise, data collection errors, lack of understanding of object intentions, etc.) as well as prevention of overfitting, i.e. learned patterns that fail to generalize to new environments~\cite{roelofs_causalagents_2022}.

Over the past few years, there has been a surge in research focused on evaluating and improving the robustness of deep learning trajectory prediction models. However, despite multiple surveys discussing trajectory prediction methods in general~\cite{huang_survey_2022,bharilya_machine_2024,Mozaffari_Deep_2022}, currently \textit{the literature lacks a detailed and comprehensive analysis of robustness strategies for trajectory prediction}. To address this gap, this paper aims to provide a structured overview of the current research on robustness in trajectory prediction. To do this, we first formalize the problem of trajectory prediction and its robustness (Section~\ref{Trajectory prediction}). We then introduce a framework that covers existing strategies for evaluating and improving robustness in trajectory prediction (Section~\ref{sec:strategies}). For the evaluation/improvement strategies, detailed insights are then provided into specific methods for making trajectory prediction models more robust (Section~\ref{Robustness approaches related to the improvement strategies}). Finally, we summarize the surveyed literature and suggest concrete directions for future research on robustness in trajectory prediction (Sections~\ref{Discussion} and~\ref{Conclusion}).

\section{Trajectory prediction and robustness} \label{Trajectory prediction}
The objective of trajectory prediction is to model the future trajectory of other road users (vehicles, pedestrians and cyclists), often referred to as agents, taking into account their past states. The state $x^{t}_{i}$ of an agent $i$ usually contains its position and potentially velocity, acceleration, and heading angle \cite{cao_advdo_2022}. Additionally, environmental contexts, such as road lines and sidewalks, are sometimes considered as well \cite{cao_advdo_2022}. The process involves capturing a sequence of observed states for each agent at a fixed time interval $\Delta t$ and generating the projected future trajectory for each agent. Some models produce a single predicted trajectory, while others generate a set of feasible trajectories along with associated confidence levels~\cite{sanchez_scenario-based_2022}. 

Formally, we will use the following description of the trajectory prediction problem~\cite{cao_advdo_2022,cao_robust_2023}. Denote by $X^{t} = (x^{t}_{1}, \ldots, x^{t}_{N})$ the joint state of the scene at time $t \leq 0$ where $N$ represents the total number of agents. Let $H$ denote the length of the observed time window, then $X = (X^{-H+1}, \ldots, X^{0})$ denote the set of states observed. Future predictions are denoted by $Y^{t} = (y^{t}_{1}, \ldots, y^{t}_{N})$ where, similarly to the past, $y^{t}_{i}$ represent the state of agent $i$ at time step $t$. The dataset consists of multiple combinations of inputs, denoted as $D = \{(X, Y),\ldots\}$. The task is to predict, using prediction model $M_\theta$ parameterized by $\theta$, the trajectory of all agents in a finite prediction horizon $T$, i.e. predict the joint trajectory $Y = (Y^{1}, \ldots, Y^{T}) \approx \hat{Y} = M_\theta(X)$.

\subsection{Standard evaluation} \label{Standard evaluation}
To train and evaluate the prediction model $M_\theta$, a train/test split of the dataset $D$ is established \cite{ian_deep_2016}. The training set is utilized to train the model by adjusting the parameters $\theta$ to optimize a loss function. The optimized model $M_{\theta^\star}$ with parameters $\theta^\star$ is evaluated on the test split of the data. Within the test set, historical states are used to generate future predictions using $M_{\theta^\star}$. Subsequently, a measure is applied to compare these predictions with the ground truth trajectories of the test set. 

\subsection{Performance measures} \label{Performance measure}
To assess the performance of the trajectory prediction model, it is crucial to quantify the disparity between the ground truth values $Y$ and the predicted values $\hat{Y}$. For this a number of measures have been developed \cite{pellegrini_youll_2009,alahi_social_2016,schmidt_lmr_2023,saadatnejad_are_2022,bahari_vehicle_2022,hallgarten_stay_2023}. When examining the measures, a critical distinction lies in whether they consider all future data points or only the endpoint. Commonly used measures include the Average Displacement Error, where the average Euclidean distance between predictions and ground truth is calculated for all time steps \cite{pellegrini_youll_2009}, and the Final Displacement Error, which considers only the endpoint \cite{alahi_social_2016}. The Miss Rate is the ratio in which the Euclidean prediction at the endpoint exceeds a defined threshold \cite{schmidt_lmr_2023}. The Collision Rate calculates the percentage of samples in which at least one collision occurs in the predicted trajectories between the candidate agent and its neighbors \cite{saadatnejad_are_2022}. The Off-Road Rate is the probability of all predicted trajectories having at least one point off-road \cite{bahari_vehicle_2022, hallgarten_stay_2023}. The Mean Inter-Endpoint Distance is the average distance from all predicted endpoints to the mean endpoint~\cite{hallgarten_stay_2023}.


\subsection{Robustness in trajectory prediction} 
Intuitively, the robustness of a prediction model $M_\theta$ requires that it produces consistent outputs $\hat{Y} = (\hat{Y}^{1}, \ldots, \hat{Y}^{T})$ even when presented with a slightly perturbed version $\tilde{X} = (\tilde{X}^{-H+1}, \ldots, \tilde{X}^{0})$ of the history $X$. Robustness against spurious features in trajectory prediction can be categorized into adversarial and natural robustness~\cite{tocchetti_i_2022,drenkow_systematic_2022}. 

Adversarial robustness involves robustness to adversarial attacks --- small, worst-case perturbations to model inputs~\cite{goodfellow_explaining_2015}. To reliably evaluate and train models that are robust to adversarial attacks, it is essential to specify an attack model, providing a precise definition of the type of attack to which the model should be resistant \cite{cao_advdo_2022}. An example of an attack model in the context of perception is presented by Eykholt et al. \cite{eykholt_robust_2018}, where black and white stickers serve as adversarial examples on road signs, achieving high targeted misclassification rates against standard-architecture road sign classifiers. 

Natural robustness of trajectory prediction implies robustness against perturbations that adhere to the physical constraints of actual vehicle driving and reflect normal driving behavior rather than adversarial maneuvers \cite{cao_advdo_2022, zhang_adversarial_2022}. If generated trajectories appear unnatural to an autonomous vehicle, they may exploit this property to predict an anomaly or an attack~\cite{zhang_adversarial_2022}. 

Moreover, ensuring the robustness of trajectory prediction model $M_\theta$ requires it to learn the true underlying relationship between the observed data $X = (X^{-H+1}, \ldots, X^{0})$ and the future predictions $Y = (Y^{1}, \ldots, Y^{T})$. Assessing its resistance to overfitting can be measured through the train/test split. The key is to determine whether the training and test data are identical and independent samples from the same data distribution \cite{liu_empirical_2022}. However, this condition is rarely satisfied in practice, resulting in distributional shifts \cite{liu_empirical_2022}. These shifts, characterized by differences between the training and test distributions, can substantially undermine the accuracy of deep learning models deployed in real-world scenarios~\cite{koh_wilds_2021}. 

\section{Strategies for evaluation \\ and improvement of robustness} \label{sec:strategies}
\begin{figure*}
    \includegraphics[width=1\linewidth]{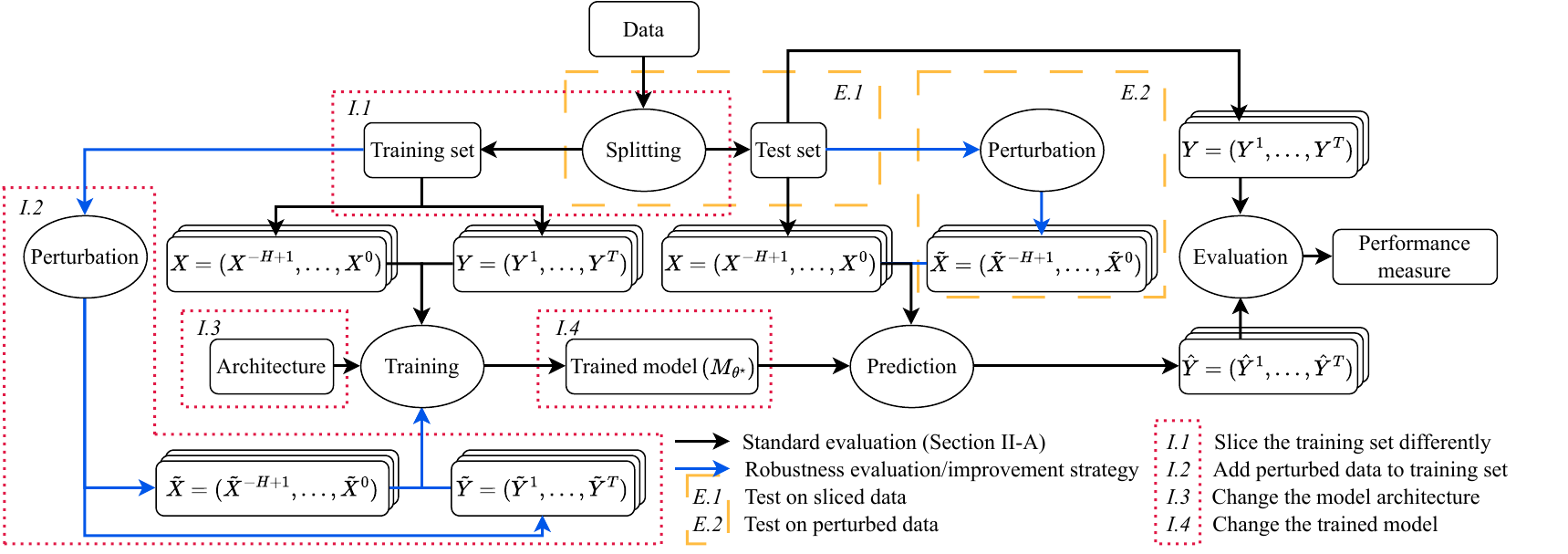}
    \caption{A general framework for training and evaluating (robust) trajectory prediction models. Robustness evaluation (\textit{E}) and improvement (\textit{I}) components are highlighted by dashed and dotted boxes respectively.}
\label{fig:overview-figure}
\end{figure*}

We propose a general framework for training and evaluating trajectory prediction models that includes robustness evaluation and improvement strategies (Figure \ref{fig:overview-figure}). This overarching framework is based on a synthesis of existing literature, incorporating approaches currently available in the field. 

\subsection{Evaluation strategies} \label{Evaluation strategies}
To assess the robustness of a trajectory prediction model in specific settings, the test data can be modified \cite{sanchez_scenario-based_2022}. We distinguish between two general strategies of evaluating robustness through data modification (Figure~\ref{fig:overview-figure}).

\subsubsection{Test on sliced data (\textit{E.1})} \label{Test on sliced data}
To assess robustness against overfitting, testing can be performed on slices \cite{chen_slice-based_2019}. This evaluation strategy involves using benign inputs from the test set under specific settings to assess the robustness of the model on crucial subsets \cite{chen_slice-based_2019} (Figure~\ref{fig:overview-figure}). The evaluation includes testing individual traffic situations sorted by factors such as safety-criticality \cite{stoler_safeshift_2024}, trajectory type \cite{sanchez_scenario-based_2022, guo_scenario_2023} or geographical location \cite{liu_survey_2024}. Whenever the model might underperform on these subsets, it poses a challenge in terms of generalizing to out-of-distribution data~\cite{liu_empirical_2022}.

\subsubsection{Test on perturbed data (\textit{E.2})} \label{Test on perturbed data }
To assess the adversarial robustness of the model, perturbations denoted as $\tilde{X}$ can be introduced into the test data \cite{zhang_adversarial_2022}. This evaluation method involves the trained trajectory prediction model $M_{\theta^\star}$ making predictions on perturbed test samples rather than benign ones (Figure~\ref{fig:overview-figure}). While models often perform well on benign inputs, they may be vulnerable to adversarial attacks and perturbations that manipulate the input, resulting in incorrect outputs \cite{madry_towards_2019}. This strategy provides insights into the robustness of the model on unseen data that have been slightly altered, offering a realistic assessment of its resilience to perturbations \cite{goodfellow_explaining_2015}. However, a notable limitation is the reliance on simulating realistic agent behavior, which can introduce a simulation-to-real gap~\cite{stoler_safeshift_2024}.

\subsection{Robustness improvement strategies } \label{Improvement strategies for robustness in trajectory prediction}
Based on the performance measure results obtained from both standard and robustness evaluations, the selection of an improvement strategy becomes crucial. Improvements in robustness can be achieved via four general strategies (Figure~\ref{fig:overview-figure}). 

\subsubsection{Slice the training set differently (\textit{I.1})} \label{Slice the training set differently }
This strategy involves modifying the training set to prevent the model from overfitting. The implementation of this approach occurs during the preparation of the train/test split (Figure~\ref{fig:overview-figure}). It includes two key steps: firstly, mining specific scenarios from the dataset, and secondly, designing a remediation strategy to mitigate the impact of distribution shift \cite{stoler_safeshift_2024}. The objective is to emphasize the overall quality of the model while improving slice-specific measures \cite{chen_slice-based_2019}. However, extracting valuable scenes poses two main challenges: first, most data is not interesting from a scenario/safety perspective, and second, extensive data processing is required, impeding the utilization of large real-world datasets~\cite{guo_scenario_2023}.

\subsubsection{Add perturbed data to training set (\textit{I.2})} \label{Introduction of perturbed data}
This strategy involves modifying the training set by introducing perturbations to enhance resistance against spurious features. It includes modifying the training data or augmenting it with perturbations (Figure~\ref{fig:overview-figure}). These perturbations can be incorporated into the historical states $X$ \cite{cao_advdo_2022, saadatnejad_are_2022, cao_robust_2023, sanchez_robustness_2023, jiao_tae_2022} and optionally into the future states $Y$ \cite{zamboni_pedestrian_2022, roelofs_causalagents_2022, bahari_vehicle_2022}, denoted as $\tilde{X}$ and $\tilde{Y}$. The objective is for the trained model to exhibit comparable behavior for perturbed and benign examples \cite{bai_recent_2021}. An important consideration for this strategy is that training the model on perturbations to improve its robustness often leads to degraded performance on clean, unperturbed data \cite{zhang_theoretically_2019}. The balance between benign and perturbed variants becomes crucial for safety-critical tasks that autonomous vehicles handle~\cite{cao_robust_2023}.

\subsubsection{Change the model architecture (\textit{I.3})} \label{Change the model architecture}
This strategy involves modifying the model architecture, focusing on the building blocks of the deep learning model (Figure~\ref{fig:overview-figure}). The building blocks of the prediction model are designed to extract features from historical states, aiming to learn the motion dynamics of each agent and the social interactions between agents \cite{cao_advdo_2022}. The model must describe how the past states of the agent influence its future movements or decisions, as well as how the state of one agent can influence the states or actions of other agents \cite{cao_advdo_2022}. Considering the data-driven approach employed during training for real-world scenarios, it becomes apparent that prediction model can face challenges in extracting features from unseen domains \cite{vishnu_improving_2023}. To address this issue, modifying the model architecture can be essential in enhancing resistance to spurious features \cite{vishnu_improving_2023} and mitigating model overfitting \cite{ye_improving_2023}. An effective method for obtaining a deeper understanding of how these building blocks affect robustness is through ablation studies~\cite{meyes_ablation_2019}.

\subsubsection{Change the trained model (\textit{I.4})} \label{Change the trained model}
This strategy involves the refinement of the trained model, with the modification taking place post-training (Figure \ref{fig:overview-figure}). Existing literature suggests that employing techniques such as model compression can enhance generalization performance on previously unseen data \cite{kuhn_robustness_2021}. A method used for this purpose is pruning, which optimizes the model by systematically removing parameters \cite{blalock_what_2020}. The prunability of a model serves as a crucial indicator of its generalization capabilities. The greater the proportion of parameters that can be pruned, the more effectively the model is expected to generalize~\cite{kuhn_robustness_2021}.

\section{Methods for evaluation\\and improvement of robustness} \label{Robustness approaches related to the improvement strategies}
For robustness evaluation/improvement strategies \textit{E.1}, \textit{E.2}, \textit{I.1}, \textit{I.2}, and \textit{I.3} (Figure~\ref{fig:overview-figure}, Section~\ref{sec:strategies}), this section describes concrete methods. At the time of writing, no papers explicitly used strategy I.4 to enhance the robustness of trajectory prediction models. 

\subsection{Slicing techniques (\textit{E.1}, \textit{I.1})} \label{Slicing techniques}
The utilization of conventional real-world data for training trajectory prediction models is attractive as it captures the authentic behaviors of road users \cite{stoler_safeshift_2024}. However, a challenge arises during data analysis: data distribution often does not favor varied scenes \cite{ivanovic_trajdata_2023}, safety-critical incidents \cite{ding_survey_2023} and geographical diversity \cite{liu_survey_2024}, a phenomenon also known as the ''Curse of Rarity'' \cite{liu_curse_2022, stoler_safeshift_2024}. The data typically contains minimal information about rare events, and acquiring sufficient information requires a substantial increase in the volume of data \cite{liu_curse_2022}. Slicing the data can be a useful method to extract valuable scenes from the dataset. These scenes are then segregated into a separate set, serving as a representation of valuable and rare, long-tailed events \cite{stoler_safeshift_2024}. After these scenes are identified, the model is trained using a remediation strategy \cite{stoler_safeshift_2024}. This process optimizes the benefits of existing data rather than necessitating the generation of new scenarios \cite{stoler_safeshift_2024}. The data can be sliced using a search that is scenario or location specific.

\subsubsection{Scenario importance sampling} \label{Scenario importance sampling}
To operate safely, an automated vehicle must anticipate the evolving environment around it. To achieve this, it is crucial to determine the most suitable prediction models for various situations \cite{sanchez_scenario-based_2022}. Addressing this challenge involves employing scenario extraction, which requires a substantial amount of data to identify common and uncommon scenarios \cite{guo_scenario_2023}. The initial step involves the predefinition of tags describing road user activities and their interactions with each other and the environment \cite{de_gelder_real-world_2020, guo_scenario_2023, sanchez_scenario-based_2022}. Subsequently, scenarios are extracted by searching for combinations of these predefined tags~\cite{de_gelder_real-world_2020, guo_scenario_2023}. 

To enhance the effectiveness of the sampling process, a more focused search can prove beneficial. An approach involves identifying and studying the most safety-critical scenarios within a given dataset \cite{stoler_safeshift_2024}. The relevance of safety warrants emphasizing scenarios in which observed agents exhibit safety-critical or near-safety-critical behavior, as well as instances in which agents avoid such critical situations through subtle and proactive maneuvers. Sampling these scenarios can be accomplished by assigning scores based on their individual state and interaction features. Individual states are derived from the relative positional data of a trajectory, including speed, acceleration, and jerk. The interaction features utilize surrogate safety metrics such as time headway, time-to-collision, and deceleration rate to avoid a crash~\cite{glasmacher_automated_2022,vogel_comparison_2003}.

\subsubsection{Geographic importance sampling} \label{Geographic importance sampling}
Datasets can be captured in various locations worldwide. The dominance of the USA, Western Europe, and East Asia leads to a bias where autonomous driving systems are overfitted to environmental conditions typical of these regions \cite{liu_survey_2024}. To address this concern, one viable approach involves the development of region-specific models that are trained on data specific to the targeted operational area. Slicing the data according to geographic regions can enhance the model's ability to comprehend and adapt to the unique characteristics of specific locations.

\subsection{Perturbation techniques (\textit{E.2}, \textit{I.2})} \label{Perturbation techniques}
As previously mentioned, an method to evaluate and improve model robustness against unseen situations involves collecting more data that includes the missing safety-critical scenarios, which is costly, time-consuming, and lacks scalability \cite{sanchez_robustness_2023}. The necessity for additional data arises due to the prevalent issue of current prediction models being frequently overfitted on limited patterns within datasets and not capturing the long-tailed distribution of driving scenarios and behaviors \cite{jiao_semi-supervised_2023}. To overcome this overfitting issue, a common practice is to use augmentation techniques to enrich existing datasets \cite{shorten_survey_2019}. Another method involves partially perturbing the training data. These perturbations can take two primary forms: complementing perturbations, designed to enrich recorded data with previously unseen situations, and disruptive perturbations, which simulate complications like sensor noise \cite{sanchez_robustness_2023}. For trajectory prediction techniques, these perturbations involve the removal of agents \cite{roelofs_causalagents_2022, sanchez_robustness_2023}, geometric transformations \cite{zamboni_pedestrian_2022}, adversarial attacks/training \cite{cao_advdo_2022, jiao_semi-supervised_2023, zhang_adversarial_2022, saadatnejad_are_2022}, and contextual map perturbations~\cite{bahari_vehicle_2022, mi_hdmapgen_2021}. 

\subsubsection{Removal of agents} \label{Removal of agents}
This method involves enhancing the data by removing historical tracked locations of agents to prevent the model from relying solely on extrapolating historical data \cite{hallgarten_stay_2023}. Two options can be considered for this task. The initial option is to completely remove historical information from a selected group of agents \cite{hallgarten_stay_2023}, optionally based on their social interactions \cite{roelofs_causalagents_2022}. Understanding how agents influence each other within the scene and labeling each agent according to its influence on interaction can be crucial for this purpose \cite{roelofs_causalagents_2022}. This can reveal whether a given agent is non-causal (an agent whose deletion does not alter the ground truth trajectory of a given target agent), causal (an agent capable of influencing the target trajectory of the agent), and static (an agent that remain motionless in the scene)~\cite{roelofs_causalagents_2022}.

Another option involves partially removing a selection of historical positions by employing methods such as dropouts \cite{song_learning_2022} or late detection \cite{sanchez_robustness_2023}. Using dropouts simulates perceptual loss, reflecting situations in the real world where the target might be lost in some timestamps or not tracked for a sufficient duration \cite{song_learning_2022}. Late detection results in no historical data to exploit, considering only the most recent observation \cite{sanchez_robustness_2023}. The significance of this improvement method lies in the need for the prediction model to handle imperfect trajectories robustly, rather than being constrained by fixed-duration tracking inputs~\cite{song_learning_2022}.

\subsubsection{Geometric transformations} \label{Geometric transformations}
Geometric transformations augment both input and ground truth labels. Rotating and mirroring data are methods commonly employed in various applications, such as segmentation problems \cite{ronneberger_u-net_2015}. These techniques can also be applied to trajectory prediction data. When each trajectory is subjected to random rotation, the directional bias can be reduced \cite{scholler_what_2020} and the network can be trained to be rotation-invariant \cite{zamboni_pedestrian_2022}. Mirroring involves flipping the trajectory over the x-axis or y-axis with a certain probability. When no rotation is applied, mirroring could potentially augment the effects of random rotations \cite{zamboni_pedestrian_2022}. Another geometric transformation method involves adding positional noise (typically Gaussian) to every point in the trajectory . This approach is intended to enhance the robustness of the network to minor perturbations and inaccuracies \cite{zamboni_pedestrian_2022}. Similarly, adding noise to the heading angle can enhance robustness against noisy heading angle measurements caused by faulty sensors or adverse weather conditions~\cite{sanchez_robustness_2023}.

\subsubsection{Adversarial attacks/training} \label{Adverserial attack} 
In this class of methods, an adversary can inject any perturbation within some limit, called the attack model, into historical trajectories \cite{jiao_semi-supervised_2023, cao_advdo_2022}. These perturbations have the potential to significantly misguide the prediction of future trajectories, leading to unsafe planning \cite{cao_robust_2023}. Effectively training the model on adversarial attacks becomes crucial for improving the safety of autonomous vehicles in uncertain scenarios~\cite{cao_robust_2023}.

The primary objective of such attacks is to manipulate predictions at each time step or partially, with the intention of causing the autonomous vehicle plan to execute unsafe driving behaviors \cite{cao_advdo_2022}. The technical challenges lie in devising an attack framework where the threat model synthesizes adversarial trajectories feasible subject to the physical constraints of the real vehicle and closely resembling nominal trajectories \cite{cao_advdo_2022, jiao_semi-supervised_2023}. Nominal trajectories are particularly important, as a significant perturbation in the trajectory history can obfuscate whether changes in future predictions result from the vulnerability of the prediction model or more fundamental perturbations to the meaning of the history \cite{cao_advdo_2022}. Once the threat model is defined, adversarial training can be applied to the trajectory models to fortify their resilience against such attacks~\cite{madry_towards_2019}.

The challenges in devising effective adversarial attacks come from the fact that historical trajectories are often generated by upstream tracking pipelines and sparsely queried using discrete time intervals due to computational limitations. On the one hand, due to the sparse sampling of these trajectories, correctly estimating the parameters of the dynamics becomes challenging, making it difficult to determine the realism of a trajectory \cite{cao_advdo_2022}. Reconstructing a realistic dense trajectory from a sampled trajectory in the dataset can mitigate this problem~\cite{cao_advdo_2022}. On the other hand, adversarial trajectories often exhibit changes in vehicle velocity or acceleration as a characteristic. A practical solution to mitigate the issue of unstable velocities or acceleration involves the application of trajectory smoothing. This approach proves beneficial in rendering the perturbation more realistic and aligning it with real-world scenarios \cite{zhang_adversarial_2022}. Both approaches ensure that adversarial trajectories are physically feasible to reproduce in the context of driving a real car. Therefore, adversarial attacks can be viewed as a method to discover the most realistic yet worst cases for the prediction model~\cite{zhang_adversarial_2022}.

\subsubsection{Contextual map perturbations} \label{Contextual map perturbations}
Recent advances in trajectory prediction models have incorporated contextual map information \cite{zheng_robustness_2023}. The inclusion of this map helps prediction algorithms recognize specific scenarios, which contributes to the improved accuracy of predictions \cite{bahari_vehicle_2022}. However, the exclusion of map information allows for the analysis of scenarios in which the road information is uncertain \cite{sanchez_robustness_2023}. Furthermore, prediction models are susceptible to a specific type of attack targeting context maps \cite{zheng_robustness_2023}. These map-based attacks involve the introduction of adversarial perturbations to the original map encoding \cite{zheng_robustness_2023}. To ensure that changes in adversarial maps are imperceptible, the generated adversarial perturbations need to be constrained~\cite{zheng_robustness_2023}. 

Another map-based method employs a generative model for adversarial scene generation. In this approach, the generative model introduces realistic modifications to induce failures in prediction models when presented with an observed scene \cite{bahari_vehicle_2022}. The generated scenarios must adhere to physical constraints and feasibility; otherwise, they cannot accurately represent potential real-world cases and can be subject to anomaly detection. Training the model on these attacks proves beneficial in improving robustness against contextual map perturbations~\cite{sanchez_robustness_2023, bahari_vehicle_2022}.

\subsection{Model architecture (\textit{I.3})} \label{Model architecture}
Modifying the model architecture can help improve various aspects of robustness in trajectory prediction. Prediction models typically consist of multiple building blocks that can be altered or expanded to improve predictive capabilities. These alterations can include a defense mechanism against perturbations \cite{zheng_robustness_2023}, scene extrapolation \cite{ye_improving_2023, hallgarten_stay_2023, wang_bridging_2023}, and comprehension of social interactions~\cite{vishnu_improving_2023, sun_recursive_2020, zhao_multi-agent_2019, wang_jointly_2022, yuan_agentformer_2021, salzmann_trajectron_2020, zhu_robust_2020, mangalam_it_2020}.

\subsubsection{Defense mechanism} \label{Defense mechanism}
Map-based attacks can significantly impact trajectory prediction models. Rather than training models on these perturbations, \cite{zheng_robustness_2023} advises to employ a defense mechanism. The defense strategy depends on how the trajectory prediction model extracts features from the map. Whether using images or node representation, defense mechanisms such as convolution-based filters or smoothers can be applied to resist these attacks~\cite{zheng_robustness_2023}. 

\newcommand*\rot[1]{\hbox to 1em{\hss\rotatebox[origin=br]{-40}{#1}}}

\begin{table*}
\centering
\caption{Surveyed papers and their utilization of performance measures (Section \ref{Performance measure}), robustness strategies (Section \ref{sec:strategies}), and robustness methods (Section \ref{Robustness approaches related to the improvement strategies})}
\begin{tabular}{l*{17}{|c}}
\multicolumn{1}{c}{\textbf{Performance measures}} & \multicolumn{1}{c}{\textbf{\rot{Zamboni et al. (2020) \cite{zamboni_pedestrian_2022}}}} & \multicolumn{1}{c}{\textbf{\rot{Wang et al. (2021) \cite{wang_bridging_2023}}}} & \multicolumn{1}{c}{\textbf{\rot{Sánchez et al. (2022) \cite{sanchez_scenario-based_2022}}}} & \multicolumn{1}{c}{\textbf{\rot{Cao et al. (2022) \cite{cao_robust_2023}}}} & \multicolumn{1}{c}{\textbf{\rot{Saadatnejad et al. (2022) \cite{saadatnejad_are_2022}}}}& \multicolumn{1}{c}{\textbf{\rot{Bahari et al. (2022) \cite{bahari_vehicle_2022}}}} & \multicolumn{1}{c}{\textbf{\rot{Cao et al. (2022) \cite{cao_advdo_2022}}}} & \multicolumn{1}{c}{\textbf{\rot{Roelofs et al. (2022) \cite{roelofs_causalagents_2022}}}}& \multicolumn{1}{c}{\textbf{\rot{Zhang et al. (2022) \cite{zhang_adversarial_2022}}}} & \multicolumn{1}{c}{\textbf{\rot{Stoler et al. (2023) \cite{stoler_safeshift_2024}}}}  &  \multicolumn{1}{c}{\textbf{\rot{Sánchez et al. (2023) \cite{sanchez_robustness_2023}}}} &  \multicolumn{1}{c}{\textbf{\rot{Zheng et al. (2023) \cite{zheng_robustness_2023}}}} & \multicolumn{1}{c}{\textbf{\rot{Jiao et al. (2023) \cite{jiao_semi-supervised_2023}}}} & \multicolumn{1}{c}{\textbf{\rot{Tan et al. (2023) \cite{tan_targeted_2022}}}} & \multicolumn{1}{c}{\textbf{\rot{Ye et al. (2023) \cite{ye_improving_2023}}}} & \multicolumn{1}{c}{\textbf{\rot{Vishnu et al. (2023) \cite{vishnu_improving_2023}}}} & \multicolumn{1}{c}{\textbf{\rot{Hallgarten et al. (2023) \cite{hallgarten_stay_2023}}}}\\  
\toprule
Average Displacement Error & \checkmark & \checkmark & \checkmark & \checkmark & \checkmark &   & \checkmark & \checkmark & \checkmark & \checkmark & \checkmark & \checkmark & \checkmark & \checkmark & \checkmark & \checkmark & \checkmark\\
Final Displacement Error & \checkmark & \checkmark & \checkmark & \checkmark & \checkmark &   & \checkmark &  & \checkmark & \checkmark &  & \checkmark &  &  & & \checkmark & \checkmark\\  
Miss Rate & & & & &  &    & \checkmark &  &  &  &  &  &  & & \checkmark & & \checkmark\\ 
Collision Rate & & & & & \checkmark &  &  &  &  & \checkmark &  &  &  &  & & &\\  
Off-Road Rate & & & & &  & \checkmark  & \checkmark &  &  &  &  &  & & & & & \checkmark\\ 
Mean Inter Endpoint Distance & & & & & &   &  &  &  &  &  &  & & & & & \checkmark\\
\midrule
\multicolumn{1}{c}{\textbf{Robustness strategies}}  \\
\midrule
Test on sliced data (\textit{E.1}) & & & \checkmark & & &  &  &  &  & \checkmark &  & & & & & \checkmark & \\
Test on perturbed data (\textit{E.2}) & & & & \checkmark & \checkmark & \checkmark  & \checkmark & \checkmark & \checkmark &  & \checkmark & \checkmark & \checkmark & \checkmark & & & \checkmark\\
\midrule
Slice the training set differently (\textit{I.1}) & & & & & &  &  &  &  & \checkmark &  &  & &  & & &\\
Add perturbed data (\textit{I.2}) & \checkmark &  & & \checkmark & \checkmark & \checkmark  & \checkmark & \checkmark & \checkmark &  & \checkmark &  & \checkmark &  &  & & \checkmark\\
Change the model architecture (\textit{I.3}) & \checkmark & \checkmark & &  & &  &  &  &  &  &  & \checkmark &  &  & \checkmark & \checkmark & \checkmark\\
Change the trained model (\textit{I.4}) & & & & &  &  &  &  &  &  &  & & & & & &\\
\midrule
\multicolumn{1}{c}{\textbf{Robustness methods}}  \\
\midrule
Scenario importance sampling & & & \checkmark &  & &  &  &  &  & \checkmark &  &  &  & & & \checkmark &\\  
Geographic importance sampling & & & &  & &  &  &  &  &  &  &  &  & & & \checkmark &\\ 
Removal of agents & & & & &  &  &  & \checkmark &  &  & \checkmark &  &  & & & & \checkmark\\ 
Geometric transformation & \checkmark & & & &  & &  & &  &  & \checkmark &  &   & & & &\\ 
Adversarial attacks/training & & & & \checkmark & \checkmark &   & \checkmark &  & \checkmark &  &  & & \checkmark  & \checkmark & & &\\ 
Contextual map perturbations & & & & &  & \checkmark &  &  &  &  & \checkmark & \checkmark  &  & & & & \checkmark\\
Defense mechanism & & & &  &  &  &  &  &  &  &  & \checkmark  & & & & &\\ 
Scene extrapolation & & \checkmark & & &  & &  &  &  &  &  &  &  & & \checkmark  & & \checkmark\\ 
Social interaction & \checkmark & & & &  & &  &  &  &  &  &  & & & & \checkmark &\\ 
\bottomrule
\end{tabular}
\label{tab:paper_comparison}
\end{table*}

\subsubsection{Scene extrapolation} \label{Scene extrapolation}
The diverse geometry and topology of map elements across different scenes pose a challenge for trajectory prediction models. Current works often neglect domain generalization, and their performance tends to degrade when transferred to a different dataset or scenario \cite{wang_bridging_2023}. Extracting domain-independent features, such as motion patterns and social interactions, rather than memorizing domain-specific map features, is beneficial for enhancing robustness against domain shift problems \cite{ye_improving_2023}. Utilizing Frenet-based coordinate systems can mitigate trajectory coordinate variations across domains by exploiting local coordinates relative to lane centerlines \cite{ye_improving_2023}. Another issue related to scene extrapolation is the significant impact of captured speed differences \cite{wang_bridging_2023}. Extreme speed cases, such as very low or high speeds, are infrequent in the training data, causing the model to struggle when tested under such conditions. Including a velocity refinement module in the model can help address this challenge.
  
\subsubsection{Social interaction} \label{Social interaction}
Vehicle trajectories encapsulate detailed semantic and behavioral information, describing the actions and intentions of road users. Among these, the understanding of social behavior and learning interactions between individuals are core challenges in trajectory prediction. Using social pooling \cite{mangalam_it_2020}, global scene information \cite{salzmann_trajectron_2020}, and spatio-temporal features \cite{zhu_robust_2020,zhao_multi-agent_2019,vishnu_improving_2023} can help models capture interaction between road users.

\section{Discussion} \label{Discussion}
Based on the survey, we identified 17 research papers that directly address the topics of evaluating and/or improving the robustness of trajectory prediction (Table~\ref{tab:paper_comparison}). From this selection, we identified four general developments. 

\subsection{Analysis of recent developments}
First, training and testing models on perturbations is the most common strategy to robustness in trajectory prediction. Perturbations are valuable tools for evaluating the performance of trajectory prediction models because they can be easily constructed and applied to discrete input data. This makes it easier to assess the robustness of the model and identify potential weaknesses.

Second, the performance in benign situations tends to degrade after including perturbations in model training \cite{jiao_semi-supervised_2023,zhang_adversarial_2022,sanchez_robustness_2023,roelofs_causalagents_2022,cao_advdo_2022}. Therefore, it is crucial to consistently evaluate the retrained model in benign situations to ascertain if it maintains the same performance as before \cite{sanchez_robustness_2023}; mixing of perturbed scenarios with benign ones during training can help mitigate this issue~\cite{zhang_mixup_2018,jiao_semi-supervised_2023}.

Third, we found that papers related to enhancing robustness in trajectory prediction often overlook the aspect of modifying the trained model (Table~\ref{tab:paper_comparison}). This gap is evident not only in the discussed papers but also appears to be a general trend within the trajectory prediction literature. 

Fourth, most existing work uses simplistic performance measures such as Average Displacement Error and Final Displacement Error. However, the relevance of these measures remains limited, particularly in evaluating adversarial attacks. These measures are designed for benign settings and are insufficient to demonstrate the implications for autonomous systems under attack \cite{cao_advdo_2022}. For instance, a large Average Displacement Error in prediction does not directly entail concrete consequences, such as a collision \cite{cao_advdo_2022}. Furthermore, when analyzing measures using only endpoints such as the Final Displacement Error, significant details regarding the trajectory shape, curvature, and point-to-point transitions are overlooked. Additionally, the measures only consider pointwise predictions whereas distributions over trajectory predictions could potentially provide a better understanding of the non-deterministic nature of human behavior~\cite{ivanovic_trajectron_2019}. 

In summary, the current research primarily focuses on evaluating and improving robustness of trajectory prediction through data perturbation using mostly ADE and FDE as performance measures (Table \ref{tab:paper_comparison}). Adversarial attacks are extensively used to generate perturbations, while adversarial training aims to enhance robustness.


\subsection{Potential research directions} \label{Potential Research Directions}

Following a comprehensive review of evaluation and enhancement strategies, we identified five key research gaps that we believe should be addressed in future literature on robustness of trajectory prediction.

First, upon analyzing modifications that can enhance the robustness of trajectory prediction methods, it becomes apparent that there is a lack of in-depth analysis regarding the strategy of modifying the trained trajectory prediction model. Research focused on evaluating the robustness of convolutional neural networks has demonstrated that post-training techniques, such as pruning, can significantly improve model robustness \cite{kuhn_robustness_2021}. Further research on this topic may reveal methods for enhancing robustness of trajectory prediction models.

Second, the papers mainly focus on enhancing robustness based on a chosen measure while evaluating the effect of the enhancements on the same measure. However, improving robustness for specific measures may conflict with one another. Investigating the influence of improving robustness with one measure over other measures can provide insights into how the measures correlate.

Third, after applying adversarial attacks during training, pre-processing these attacks during both training and testing phases can be essential to enhance the robustness of the model. Previous studies have indicated that managing the trade-off between clean and perturbed data \cite{jiao_semi-supervised_2023}, exploring defense strategies \cite{zheng_robustness_2023}, and smoothing the train and test data \cite{jiao_semi-supervised_2023,zhang_adversarial_2022} can significantly influence the resilience of the model against adversarial examples. A comprehensive examination of all pre-processing possibilities for adversarial data in trajectory prediction models can be conducted.

Fourth, when analyzing papers related to adversarial training in trajectory prediction \cite{cao_advdo_2022, jiao_semi-supervised_2023, tan_targeted_2022, zhang_adversarial_2022}, the focus is usually on evaluating the performance of the same models. To assess the effectiveness of adversarial training strategies across diverse algorithms, a direct comparison involving different prediction models and datasets can be undertaken. This comparative analysis can help investigate the generalizability of adversarial training for trajectory prediction models.

Fifth, the majority of papers present only their method, and do not explore how it combines with existing robustness methods. As a result, the interactions of various robustness methods remain unclear. Designing a unified framework that allows for the combination and comparison of these methods will facilitate the exploration of their synergies and the identification of underexplored methods.


\section{Conclusion} \label{Conclusion}
This survey presents a framework that categorizes literature related to the evaluation and improvement of robustness in trajectory prediction. We found that existing literature in the field mainly focuses on several mainstream general strategies and a number of common methods, as well as shares some key limitations. In particular, we observed the widespread use of perturbations for evaluating and improving robustness, the limited adoption of post-training techniques to enhance robustness, and the lack of diversity in evaluation measures to validate prediction performance. We believe that addressing the identified challenges will advance the development of robust and reliable trajectory prediction models for autonomous vehicles.

\bibliographystyle{jabbrv_ieeetr}
\bibliography{IEEEabrv,articles}

\end{document}